\documentclass[letterpaper, 10 pt, conference]{ieeeconf}  

\IEEEoverridecommandlockouts                              

\usepackage{xspace}
\usepackage{xcolor}
\usepackage{amsfonts}
\usepackage{graphics}
\graphicspath{{figures/}}
\usepackage{epsfig}
\usepackage{mathptmx}
\usepackage{times}
\usepackage{amsmath}
\usepackage{amssymb}
\usepackage{hyperref}
\usepackage{ulem}
\usepackage{siunitx}
\usepackage{verbatim}
\usepackage{booktabs}
\usepackage{multirow}
\usepackage[font=small,labelfont=bf]{caption}
\usepackage{algorithm}
\usepackage{algorithmic}

\newcommand{\method}{\textit{DexMulti}\xspace}

\usepackage{color, soul}

\title{\LARGE \bf
Concurrent Prehensile and Nonprehensile Manipulation: \\ A Practical Approach to Multi-Stage Dexterous Tasks
}

\author{Hao Jiang \quad
Yue Wu \quad
Yue Wang \quad
Gaurav S. Sukhatme \quad
Daniel Seita}%

\begin{document}

\twocolumn[{%
\renewcommand\twocolumn[1][]{#1}%
\maketitle
\vspace{-1.5em}
\begin{center}
    \centering
    \captionsetup{type=figure}
    \includegraphics[width=1.0\textwidth]{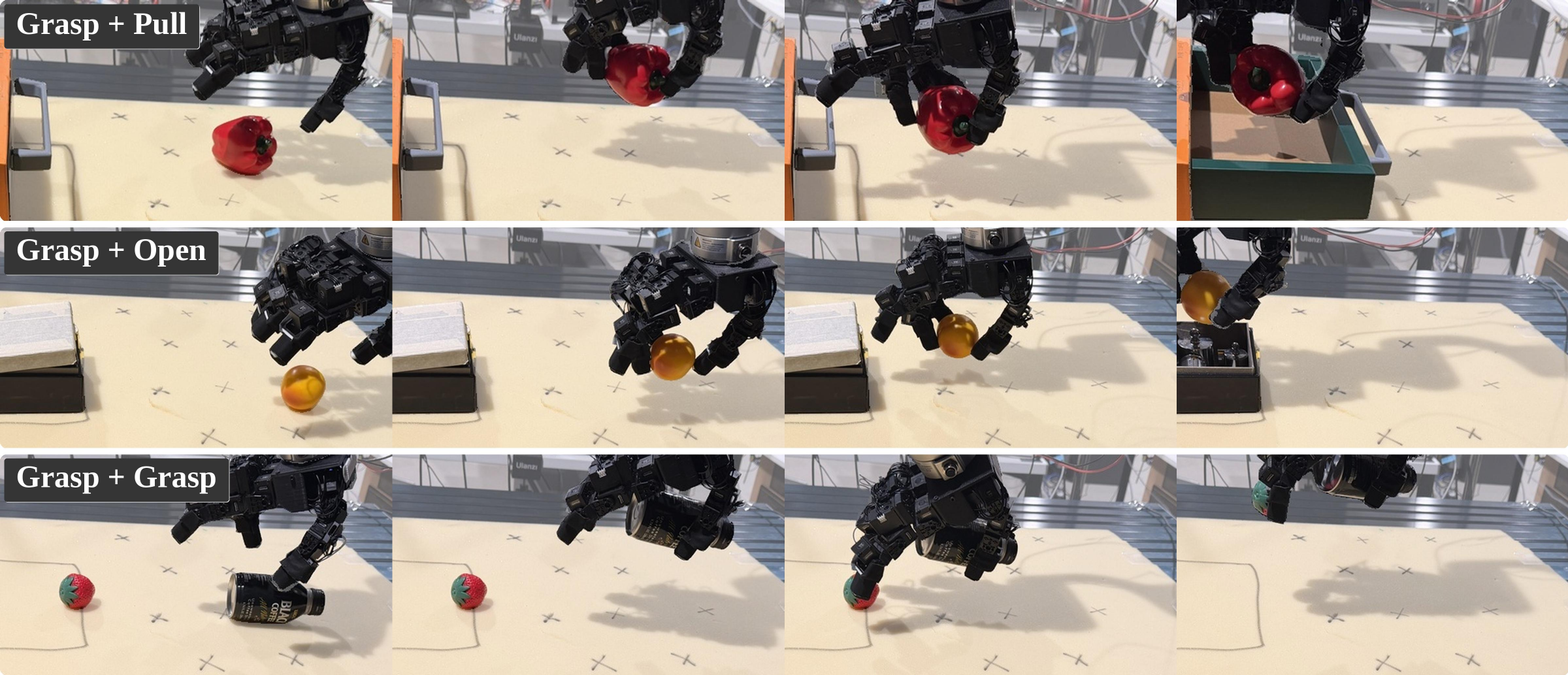}
    \vspace{-10pt}
    \captionof{figure}{
        Our method enables real-world dexterous multi-stage manipulation requiring concurrent prehensile and nonprehensile interaction. We demonstrate three challenging tasks: \textbf{(Top)} Grasp+Pull: grasping an object and pulling open a drawer while maintaining the grasp; \textbf{(Middle)} Grasp+Open: grasping an object and opening a container lid; \textbf{(Bottom)} Grasp+Grasp: sequentially grasping two objects without releasing the first. Each row shows key frames from a successful rollout, highlighting the robot's ability to stably hold one object while manipulating another. More videos and code are available on our project website \url{https://dexmulti.github.io/}.
    }
    \label{fig:teaser}
\end{center}%
}]
\begingroup
  \setlength{\skip\footins}{1pt}
  \setlength{\footnotesep}{2pt}
  \let\thefootnote\relax
  \footnotetext{All authors are with the Thomas Lord Department of Computer Science at the University of Southern California, USA. Correspondence: \texttt{hjiang86@usc.edu}.}
\endgroup

\thispagestyle{empty}
\pagestyle{empty}


\begin{abstract}
Dexterous hands enable concurrent prehensile and nonprehensile manipulation (holding one object while interacting with another), a capability essential for everyday tasks yet underexplored in robotics. 
Learning such long-horizon, contact-rich multi-stage behaviors is challenging: demonstrations are expensive to collect, and end-to-end policies require substantial data to generalize across varied object geometries and placements.
We present \method, a sample-efficient approach for real-world dexterous multi-task manipulation that decomposes demonstrations into object-centric skills with well-defined temporal boundaries.
Rather than learning monolithic policies, our method retrieves demonstrated skills based on current object geometry, aligns them to the observed object state using an uncertainty-aware estimator that tracks centroid and yaw, and executes them via a retrieve-align-execute paradigm.
We evaluate on three multi-stage tasks requiring concurrent manipulation (Grasp+Pull, Grasp+Open, and Grasp+Grasp) across two dexterous hands (Allegro and LEAP) in over 1,000 real-world trials.
Our approach achieves an average success rate of 66\% on training objects with only 3-4 demonstrations per object, outperforming diffusion policy baselines by 2-3$\times$ while requiring 5 demonstrations compared to 20-50 for comparable performance.
Results demonstrate robust generalization to held-out objects and spatial variations up to $\pm$25\,cm.
\end{abstract}

\section{Introduction}
\label{sec:intro}

Dexterous hands promise a form of manipulation that is qualitatively different from parallel-jaw grasping: they can stably hold an object while simultaneously interacting with a second object or mechanism.
This \textbf{concurrent prehensile and nonprehensile} capability underlies many everyday behaviors, such as holding a key while turning a doorknob, stabilizing a container while opening its lid, or carrying an item while pulling a drawer.
Despite this, most prior work in dexterous manipulation studies skills centered around a \textbf{single} manipulated object at a time or evaluates policies on relatively short-horizon interactions~\cite{openai2019solvingrubikscuberobot,qi2022hand, wang2023dexgraspnet,fu2025cordvip}.
As a result, it remains challenging to build systems that can robustly execute \textbf{long-horizon, contact-rich, multi-stage} tasks that require sequential interactions with multiple objects while maintaining stable grasps.

For learning-based approaches, a central difficulty is \textbf{data}.
Long-horizon dexterous demonstrations are time-consuming and physically taxing to collect, particularly when they require precise finger coordination and sustained contact.
At the same time, multi-stage tasks exhibit substantial geometric diversity: success depends on fine alignment between the hand and objects (e.g., a handle) under varying initial placements and object shapes.
End-to-end visuomotor policies, including diffusion-based imitation learning approaches, have shown impressive results when trained with sufficient and diverse demonstrations, but their performance can degrade sharply in low-data regimes and under spatial variation~\cite{chi2023diffusionpolicy,Ze2024DP3}.
This tension motivates a question that we view as fundamental for practical learning-based dexterous manipulation:

\begin{quote}
\textit{How can we learn long-horizon, contact-rich dexterous multi-task manipulation that generalizes across object instances and placements, using only a small number of real demonstrations?}
\end{quote}

In this work, we hypothesize that a key missing ingredient is an \textbf{object-centric skill abstraction}.
Rather than attempting to learn one monolithic policy that maps observations to actions over an entire episode, we decompose demonstrations into reusable object-centric skills with well-defined temporal boundaries.
This enables a retrieve-align-execute paradigm: given the current object geometry, we retrieve a relevant demonstrated skill, align it to the current object state, and execute it with a stable controller.
Crucially, this decomposition isolates the most contact-sensitive parts of the behavior into short segments that can be reliably replayed once the hand is aligned, improving generalization in low-data settings.

We instantiate this idea in \method, a real-world dexterous manipulation system for multi-stage two-object tasks.
\method combines: (i) a real-time perception pipeline that produces object-centric point clouds and interaction signals; (ii) an object-centric skill segmentation procedure that extracts \textbf{pre-manipulation} keyframes from contact-derived signals; and (iii) a retrieval-based execution mechanism that aligns and replays demonstrated skills in an object-centric frame.
Our perception and execution pipeline is designed to be robust to imperfect state estimates rather than assuming full observability.
We evaluate \method on a suite of real-world dexterous multi-task problems (Fig.~\ref{fig:teaser}), demonstrating strong performance with limited demonstrations.
Across tasks and objects, we find that object-centric skill decomposition provides a strong inductive bias and achieves robust performance with limited demonstrations, outperforming end-to-end diffusion baselines in our experiments.

The contributions of this paper include:
\begin{itemize}
    \item A formalism of \textbf{dexterous multi-task} manipulation that requires concurrent prehensile and nonprehensile interaction, a setting common in daily life but underexplored in prior dexterous manipulation research.
    \item An \textbf{object-centric skill abstraction} for long-horizon dexterous tasks, including a practical keyframe extraction procedure based on interaction signals.
    \item A retrieval-based execution framework that aligns and replays object-centric skills, improving spatial generalization and reducing demonstration requirements.
    \item Comprehensive real-world experiments across two dexterous hand platforms, including baseline comparisons, sample-efficiency analysis, and embodiment transfer.
\end{itemize}

\section{Related Work}

\subsection{Dexterous Hand Manipulation}

Dexterous manipulation with multi-fingered hands enables more complex maneuvers such as in-hand manipulation~\cite{openai2019solvingrubikscuberobot,patidar2023inhandcubereconfigurationsimplified} compared to suction cup or parallel-jaw grippers, and has been the subject of research for decades~\cite{rus1999inhanddexmanip}. One research direction studies analytical model-based approaches for dexterous grasping~\cite{li2023froggerfastrobustgrasp}. This computation may involve differentiable planners~\cite{liu2020deepdifferentiablegraspplanner} or force-closure objectives~\cite{liu2022differentiableforceclosure} to determine grasp quality. 
Such metrics have recently seen success in providing supervision for large-scale dexterous hand datasets~\cite{wang2023dexgraspnet,he2025dexgraspnet3,lum2024gripmultifingergraspevaluation}, which can facilitate downstream supervised learning approaches. 
Beyond grasping, researchers have also recently explored learning-based approaches for dexterous hand manipulation. These approaches are typically based on deep reinforcement learning in simulation with sim-to-real transfer~\cite{SOPE_2024,RobotSynesthesia2024,openai-dactyl,chen2023sequential}, or imitation learning~\cite{Ze2024DP3} from demonstrations, typically with teleoperation hardware~\cite{wang2024dexcap}. 
Our work falls under imitation learning approaches, but differs from prior work in that we systematically explore strategies for \emph{improving multi-step manipulation} involving a mix of nonprehensile~\cite{mason1999nonprehensile} and prehensile manipulation.

\subsection{Multi-Task and Concurrent Manipulation}

Dexterous hands are especially well-suited for \emph{multi-task} manipulation, due to their flexibility and degrees of freedom~\cite{chen2024task}. 
One type of multi-task manipulation involves \emph{multi-object} grasping~\cite{sun2022multi,chen2021multiobjectgraspingestimating,lu2025graspahandful}, where a hand enables efficient grasping of objects. Classical work has studied multi-object grasping through theoretical grasp analysis~\cite{yamada2015static,yu2001internalforces,yoshikawa2001powergrasp,harada1998kinematics}. 
Unlike analytic approaches, we aim to perform real-world grasping autonomously using perception data and demonstrations, and without humans inserting objects in the hand~\cite{yao2023exploiting}. 
Furthermore, unlike recent research showing promising real-world deployment of dexterous, multi-object grasping in single~\cite{li2024grasp} and sequential~\cite{he2025sequential} settings, we study more general multi-task problems beyond solely grasping. Our goal is to develop a recipe and framework for dexterous \emph{multi-step} manipulation which involves smoothly integrating both nonprehensile and prehensile manipulation, going beyond recent work that studies dexterous nonprehensile-only manipulation~\cite{li2026nonprehensile} and without relying on specialized, custom-made hands~\cite{eom2024mogrip}. 


\subsection{Sequencing and Retrieval in Multi-Task Manipulation}

Our research is also related to multi-step manipulation, which involves a robot performing multiple, clearly-defined steps. 
In closely-related work,~\cite{chen2023sequential} propose a reinforcement learning procedure for multi-step tasks in simulation with transfer to real, utilizing forward-backward fine-tuning to facilitate sequential task compatibility. Our approach is complementary, as we study multi-task imitation learning without simulation. 
Our approach is inspired by studies of design choices for data-efficient learning of multiple tasks, including decompositions that isolate an alignment phase~\cite{Dreczkowski2025thousandtasksday} and retrieval over large databases to transfer action-relevant affordances~\cite{kuang2024ramretrievalbasedaffordancetransfer}. We show how to adapt these ideas for dexterous hands. 
Furthermore, in contrast to these preceding works, our robotic hand must maintain control of an existing task while \emph{concurrently} performing the next task.

\begin{figure*}[t]
\center
\includegraphics[width=\textwidth]{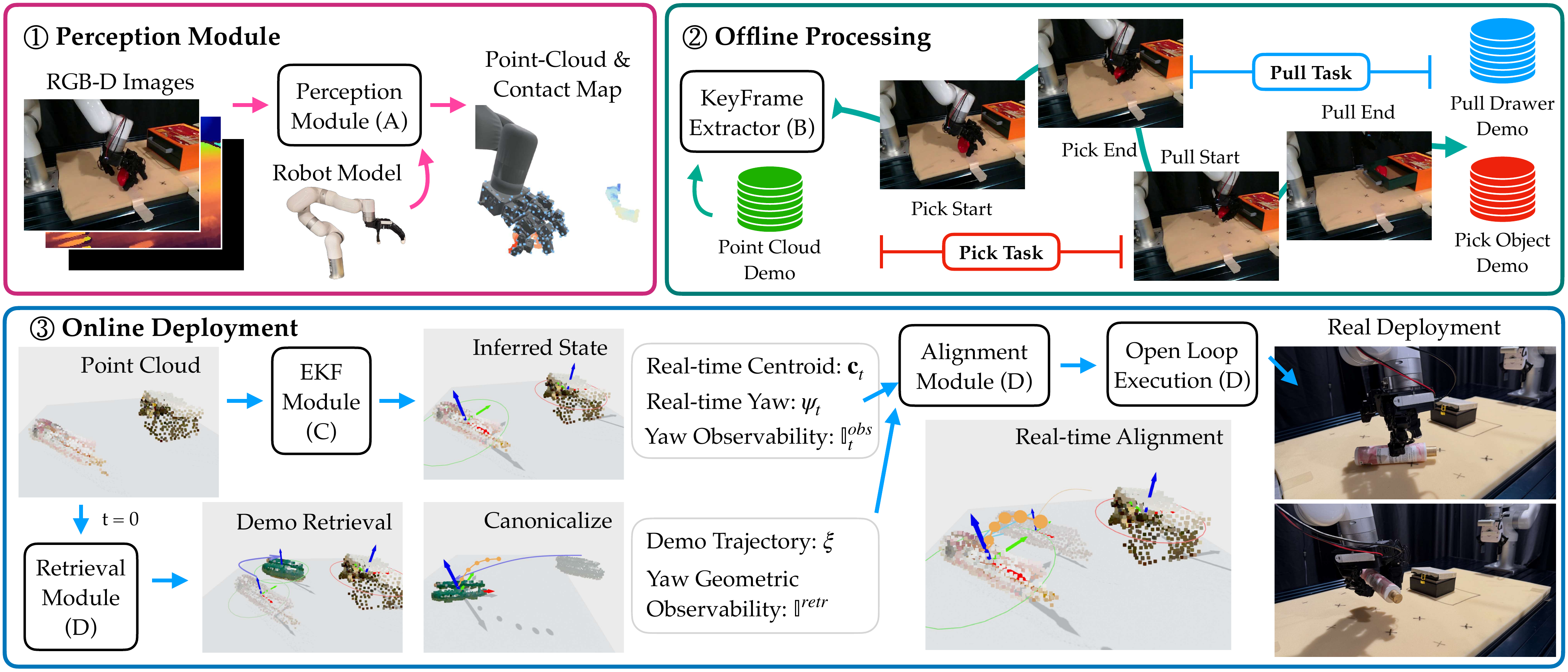}
\caption{
Overview of the proposed pipeline.
\textbf{(1) Perception:} Multi-view RGB-D with language-conditioned segmentation produces object-centric point clouds and contact maps.
\textbf{(2) Offline:} Demonstrations are segmented into object-centric skills using interaction signals and stored in canonical form.
\textbf{(3) Online:} An uncertainty-aware estimator tracks object centroid and yaw. Skills are retrieved via point-cloud matching, aligned under pose uncertainty, and executed via retrieve--align--execute.
Panels (A–D) correspond to Sec.~IV-A--D.
}
\label{fig:system}
\vspace{-10pt}
\end{figure*}

\section{Problem Statement and Assumptions}
\label{sec:problem}

We consider multi-stage dexterous manipulation tasks in which a robot arm equipped with a $d_h$-DoF multi-fingered hand must sequentially interact with multiple objects on a tabletop.
Each task is specified by a natural-language instruction $\ell$ (e.g., ``grasp the apple and open the drawer'') that determines both the relevant object categories and the order of interaction, and consists of an ordered sequence of $K$ object-centric subtasks $\tau = (s_1, s_2, \ldots, s_K)$, where each subtask $s_k$ targets a specific object $o_k$ and requires a distinct manipulation behavior (e.g., grasping, pulling, pushing).
At each timestep $t$, the robot observes multi-view RGB-D images $\{\mathbf{I}^{(v)}_t, \mathbf{D}^{(v)}_t\}_{v=1}^{V}$ from $V$ calibrated cameras and its proprioceptive state $\mathbf{q}_t = (\mathbf{q}^{arm}_t, \mathbf{q}^{hand}_t)$.
The action space is $\mathbf{a}_t = (\mathbf{a}^{arm}_t, \mathbf{a}^{hand}_t)$, where $\mathbf{a}^{arm}_t \in SE(3)$ is a target Tool Center Point (TCP) pose and $\mathbf{a}^{hand}_t \in \mathbb{R}^{d_h}$ specifies target hand joint positions.
All objects rest on a planar surface, so their poses are characterized by a 3D centroid $\mathbf{c} \in \mathbb{R}^3$ and a yaw angle $\psi \in (-\pi, \pi]$ about the vertical axis; the robot may maintain contact with a grasped object while interacting with the next.

We assume access to a small dataset of $N$ expert demonstrations $\mathcal{D} = \{\xi_1, \ldots, \xi_N\}$ collected via teleoperation, where each $\xi_i = \{(\mathbf{q}_t, \mathbf{a}_t, \mathbf{I}_t, \mathbf{D}_t)\}_{t=1}^{T_i}$ is a full episode spanning the complete multi-stage task.
Demonstrations are collected across a set of training objects with randomized placements, with only 3--4 demonstrations per object.
Our goal is to execute all $K$ subtasks in sequence given a new scene with potentially unseen object instances at arbitrary placements, using this limited demonstration budget.

\section{Proposed Method}
\label{sec:method}

\method decomposes each multi-stage task into object-centric skills that are retrieved and replayed at execution time (see Fig.~\ref{fig:system} and Alg.~\ref{alg:deploy}).
The pipeline consists of: (i) a perception module that produces object-centric point clouds from multi-view RGB-D (Sec.~\ref{ssec:perception}), (ii) an offline procedure that segments demonstrations into canonical object-centric skills (Sec.~\ref{ssec:skill_seg}), (iii) an uncertainty-aware object state estimator (Sec.~\ref{ssec:state_est}), and (iv) a retrieve-align-execute policy for deployment (Sec.~\ref{ssec:retrieval}).

\subsection{Perception}
\label{ssec:perception}

From multi-view RGB-D images $\{\mathbf{I}^{(v)}_t, \mathbf{D}^{(v)}_t\}$, we extract an object-centric point cloud $\mathbf{P}_k \subset \mathbb{R}^3$ for each task-relevant object $o_k$.
Given the language instruction $\ell$, we use Grounding DINO~\cite{liu2024grounding} to detect each object category in 2D, then SAM2~\cite{ravi2024sam2} for real-time instance segmentation and tracking.
Segmented pixels are back-projected into 3D using known camera parameters, and the resulting point clouds are denoised, clustered, and downsampled to a fixed size of 512 points via farthest point sampling (see Appendix for details).
All points are expressed in the robot base frame, and the full pipeline runs at approximately 20\,Hz.

\subsection{Object-centric Skill Decomposition}
\label{ssec:skill_seg}

Each demonstration $\xi_i$ spans the full multi-stage task.
We decompose it into $K$ object-centric skill segments $\xi_i \rightarrow (\sigma_{i,1}, \ldots, \sigma_{i,K})$, where $\sigma_{i,k}$ is the sub-trajectory corresponding to subtask $s_k$ targeting object $o_k$.

\paragraph{Segmentation via Contact Signals}
Segment boundaries are determined by hand-object interaction.
We model the dexterous hand as a point cloud obtained via forward kinematics~\cite{dexpoint} from the recorded joint states, and compute a soft contact signal measuring geometric proximity between the hand and each object's point cloud $\mathbf{P}_k$~\cite{fu2025cordvip}.
For each object $o_k$, the keyframe marking the start of $\sigma_{i,k}$ is the first timestep at which sustained hand-object contact is detected.

\paragraph{Canonical Skill Representation}
Each skill segment $\sigma_{i,k}$ is stored in a canonical object-centric frame.
Given the object point cloud $\mathbf{P}_k$ at the skill's keyframe, we compute its centroid $\mathbf{c}$ and a canonical yaw $\psi^{canon}$ from Principal Component Analysis (PCA) on the horizontal projection, and transform all points as
\begin{equation}
\mathbf{P}^{canon} = \{ \mathbf{R}_z(-\psi^{canon})(\mathbf{p}_i - \mathbf{c}) \}.
\end{equation}
All end-effector poses and hand joint commands in $\sigma_{i,k}$ are expressed in the same canonical frame, yielding a skill library that is invariant to object translation and, when yaw is observable, rotation.

\subsection{Object-centric State Estimation}
\label{ssec:state_est}

During execution, we maintain a real-time estimate of each object's centroid $\mathbf{c}_t$ and yaw $\psi_t$ using an extended Kalman filter (EKF) with constant-velocity dynamics.
Object position is measured as the centroid of the segmented point cloud; yaw is measured via PCA on the horizontal projection when the object exhibits sufficient geometric anisotropy.

\paragraph{Design Rationale}
Our centroid+yaw representation is motivated by the planar tabletop setting: 
objects resting on a flat surface have rotation constrained to the vertical axis (perpendicular to the table), as tilting is prevented by gravity and surface contact.
This reduces the state estimation burden compared to full 6D pose tracking while retaining the information needed for successful manipulation alignment.

A key design choice is explicit \textbf{yaw observability} reasoning.
The estimator computes a continuous observability score from geometric cues (e.g., point cloud anisotropy) and Kalman filter uncertainty, producing a binary yaw-observable flag $\mathbb{I}^{obs}_t$ via hysteresis thresholding.
When yaw is deemed unobservable (e.g., for round or textureless objects), orientation updates are suppressed and the task-level yaw $\psi^{task}_t$ is held constant, preventing spurious rotations. 
This contrasts with neural 6D pose trackers such as FoundationPose~\cite{foundationposewen2024}, which implicitly assume orientation is always observable; we show in Sec.~\ref{ssec:ablation} that explicit observability reasoning is critical for stable manipulation.

\subsection{Retrieve-Align-Execute}
\label{ssec:retrieval}

At execution time, each subtask $s_k$ is handled by retrieving, aligning, and replaying a matching skill from the library.

\paragraph{Retrieval}
Given the current object point cloud $\mathbf{P}^{obs}_k$, we retrieve the closest stored skill via Chamfer distance in canonical space:
\begin{equation}
i^* = \arg\min_i \; \min_{\theta \in \Theta}
\; d_{\mathrm{Chamfer}}\!\left(
\mathbf{R}_z(\theta)\mathbf{P}^{obs}_{centered},\;
\mathbf{P}^{canon}_i
\right),
\end{equation}
where $\Theta$ is a discrete set of yaw hypotheses.
When yaw is observable ($\mathbb{I}^{obs}_t = 1$), $\theta$ is fixed to $\psi^{task}_t$ and only one Chamfer distance is evaluated.
When yaw is unobservable, we minimize over $\Theta$, jointly recovering the best demonstration $i^*$ and a relative alignment yaw $\theta^*$.
For multi-stage tasks, we retrieve a single demonstration index $i^*$ using a joint score across all task-relevant objects, ensuring consistent skill pairings across subtasks.

\paragraph{Alignment}
After retrieval, the robot performs closed-loop alignment by tracking the first canonical action of the retrieved skill, transformed to the world frame using the real-time object state estimate:
\begin{equation}
\mathbf{p}^{world} = \mathbf{c}_t + \mathbf{R}_z(\psi_t)\,\mathbf{p}^{canon},
\quad
\mathbf{R}^{world} = \mathbf{R}_z(\psi_t)\,\mathbf{R}^{canon},
\end{equation}
where $\psi_t$ is selected based on both retrieval-time and real-time yaw observability (see Appendix for the full selection rule).
At each control step, we use the Pyroki optimizer~\cite{kim2025pyroki} to solve a receding-horizon motion planning problem that tracks $(\mathbf{p}^{world}, \mathbf{R}^{world})$ while enforcing collision avoidance against perception-derived obstacle representations, joint limits, and trajectory smoothness.

\paragraph{Execution}
Once alignment converges, the robot smoothly transitions via pose interpolation and then executes the remainder of the skill open-loop, replaying the canonical trajectory in the world frame using the object pose $(\mathbf{c}_{fixed}, \psi_{fixed})$ frozen at the end of alignment.
This avoids instability from continuous pose re-estimation during contact-rich manipulation.
For multi-stage tasks, this retrieve-align-execute procedure is applied sequentially to each subtask $s_k$.

\begin{algorithm}[t]
\caption{Deploy-time multi-stage retrieve--align--execute}
\label{alg:deploy}
\begin{algorithmic}[1]
\REQUIRE Task instruction $\ell$, skill library $\{\sigma_{i,k}^{canon}\}$
\STATE Parse $\ell$ to obtain subtask sequence $(s_1, \ldots, s_K)$
\STATE Detect and track objects $(o_1, \ldots, o_K)$
\STATE Observe point clouds $\{\mathbf{P}^{obs}_k\}$ for all objects
\STATE Retrieve demo $i^* \!=\! \arg\min_i \sum_k d_{\mathrm{Chamfer}}(\mathbf{P}^{obs}_k, \mathbf{P}^{canon}_{i,k})$
\FOR{$k = 1, \ldots, K$}
    \STATE \textbf{// Closed-loop alignment}
    \WHILE{not converged}
        \STATE Update state estimate $(\mathbf{c}_t, \psi_t, \mathbb{I}^{obs}_t)$ from EKF
        \STATE Transform canonical target to world frame (Eq.~3)
        \STATE Solve motion planning via Pyroki; send joint cmd
    \ENDWHILE
    \STATE Record $(\mathbf{c}_{fixed}, \psi_{fixed})$ at convergence
    \STATE \textbf{// Open-loop execution}
    \FOR{each remaining action in $\sigma_{i^*,k}^{canon}$}
        \STATE Transform to world frame using $(\mathbf{c}_{fixed}, \psi_{fixed})$
        \STATE Send arm pose and hand joint cmd
    \ENDFOR
\ENDFOR
\end{algorithmic}
\end{algorithm}

\section{Experiments}
\label{sec:experiments}

We conduct comprehensive real-world experiments to answer the following questions:

\begin{itemize}
    \item How does our method perform on diverse dexterous multi-task manipulation problems compared to learning-based baselines? (\S\ref{ssec:overall_perf})
    \item How sample-efficient is our retrieval-based approach compared to end-to-end diffusion policies? (\S\ref{ssec:sample_eff})
    \item How do state estimation and object-centric representation choices affect manipulation performance? (\S\ref{ssec:ablation})
    \item Does our method generalize across mechanically different dexterous hand embodiments? (\S\ref{ssec:embodiment})
\end{itemize}


\subsection{Experimental Setup}
\label{ssec:exp_setup}

\begin{figure}[t]
\center
\includegraphics[width=0.95\linewidth]{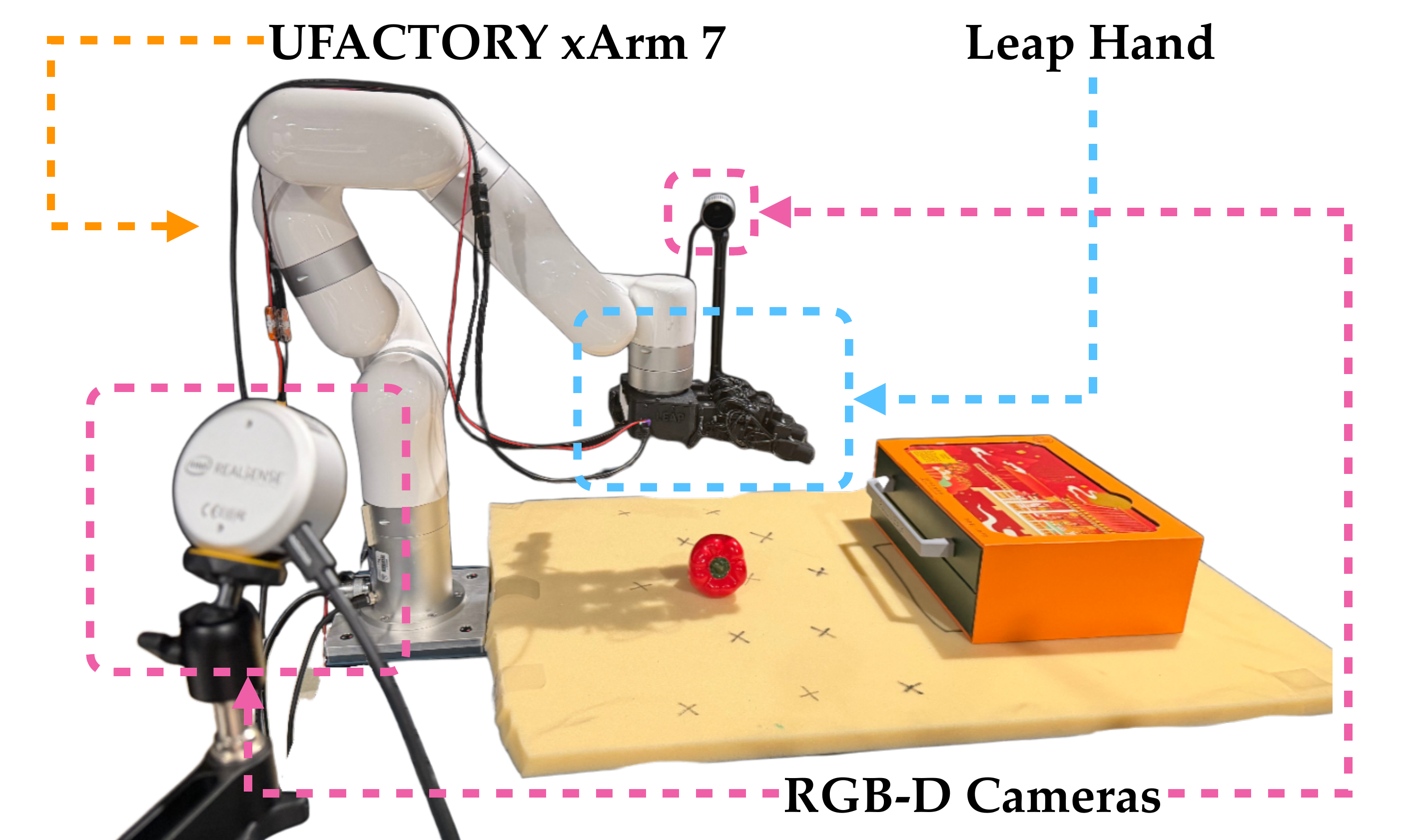}
\caption{
  Experimental setup. We use an xArm 7 robot arm with a dexterous hand (16-DoF LEAP or Allegro Hand) and two Intel RealSense L515 RGB-D cameras for multi-view perception.
}
\label{fig:phys_setup}
\vspace{-10pt}
\end{figure}

\paragraph{Robot Platform and Control}
All experiments are conducted on a 7-DoF xArm7 robotic manipulator equipped with a dexterous hand (see Fig.~\ref{fig:phys_setup}).
We evaluate two end-effectors: a 16-DoF Allegro Hand and a LEAP Hand~\cite{shaw2023leaphand}, enabling experiments across different dexterous embodiments.
The workspace is observed by two Intel RealSense L515 RGB-D cameras at complementary viewpoints, providing multi-view input to the perception pipeline described in Sec.~\ref{ssec:perception}.
Both arm and hand commands are issued at 10\,Hz; additional details on control frequencies, velocity limits, and command interfaces are provided in the Appendix.

\paragraph{Task Suite}
We evaluate on a suite of real-world dexterous multi-task manipulation problems requiring sequential interaction with multiple objects.
Each task consists of two object-centric subtasks executed in sequence:
\begin{itemize}
    \item \textbf{Grasp+Pull:} grasp an object from the table, pull open a drawer by its handle, and place the object in the drawer.
    \item \textbf{Grasp+Open:} grasp an object, push open a container lid, and place the object inside.
    \item \textbf{Grasp+Grasp:} sequentially grasp two objects without releasing the first, similar to~\cite{he2025sequential}.
\end{itemize}
Critically, all tasks require concurrent manipulation: the robot must maintain a stable grasp on the first object while performing the second subtask.
This tests a key capability that distinguishes dexterous hands from parallel-jaw grippers.
Across all tasks, object shapes, sizes, and initial placements are randomized (within a $\pm$25\,cm workspace region) to evaluate robustness and spatial generalization.

\begin{figure}[t]
\centering
\includegraphics[width=0.95\linewidth]{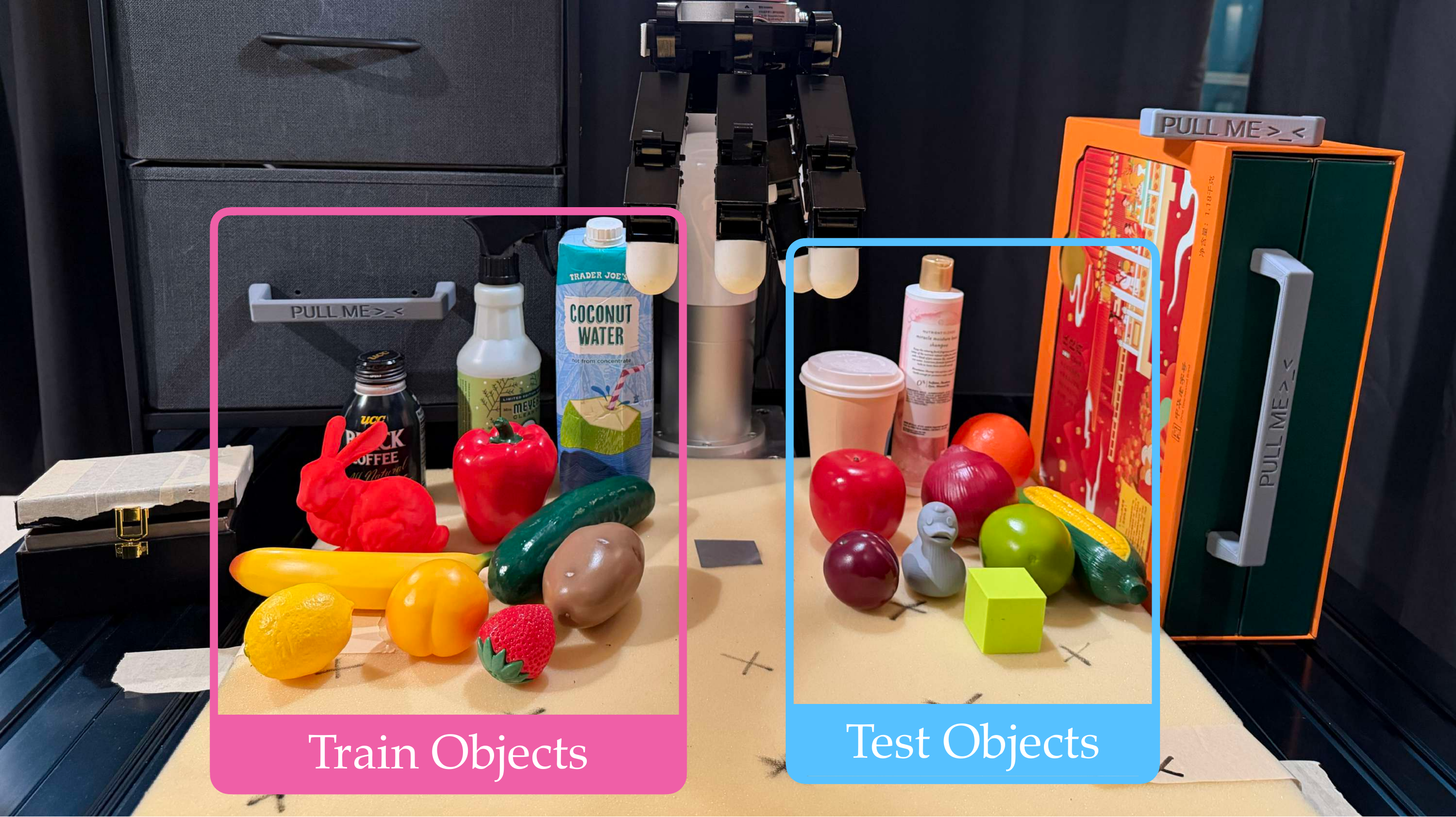}
\caption{
Objects used in our experiments.
\textbf{Left:} training objects for demonstration collection.
\textbf{Right:} held-out test objects used exclusively for evaluation, and unseen during training.
The set spans a diverse range of geometries, sizes, and interaction affordances.
}
\label{fig:objects}
\vspace{-10pt}
\end{figure}

Fig.~\ref{fig:objects} shows all training and test objects.

\paragraph{Demonstrations}
Expert demonstrations are collected via the DexCap teleoperation system~\cite{wang2024dexcap}.
For each task, we collect 3-4 demonstrations per object across 13 distinct objects, yielding approximately 40 demonstrations per task.
To evaluate sample efficiency, we additionally collect up to 50 demonstrations for selected object-task combinations.
All demonstrations use randomized object placements.
Further details on recording format and data contents are in the Appendix.

\paragraph{Baselines}
We test three learning-based baselines:
\begin{itemize}
    \item \textbf{Image-based DP:} a diffusion policy trained on RGB observations~\cite{chi2023diffusionpolicy}.
    \item \textbf{DP3 (raw point cloud):} a diffusion policy conditioned on raw scene point clouds~\cite{Ze2024DP3}.
    \item \textbf{Object-centric point cloud diffusion:} a diffusion policy conditioned on object-centric point clouds, similar in spirit to CordViP-style formulations~\cite{fu2025cordvip}.
\end{itemize}


The object-centric baseline extends DP3~\cite{Ze2024DP3} with per-object segmented point clouds following CordViP~\cite{fu2025cordvip}. For each object, we compute a per-point contact map and concatenate it as an additional channel. Each object is encoded via PointNet~\cite{qi2017pointnet} and fused with proprioceptive state; the diffusion backbone is shared with the other baselines. This is our strongest baseline; details are in the Appendix.

\subsection{Overall Task Performance}
\label{ssec:overall_perf}

We evaluate overall task success across the suite of dexterous multi-task manipulation problems, comparing our method against learning-based diffusion policies.
We evaluate using \textbf{training objects} (seen during demonstration collection) and \textbf{test objects} (held out during training; see Fig.~\ref{fig:objects}).
Each method is evaluated over multiple trials per task-object pair with randomized initial object placements; the number of trials varies by task as shown in Table~\ref{tab:overall_performance}.

\paragraph{Metrics}
We report task success rate, defined as completing \textbf{both} subtasks in sequence without external intervention.
A trial is considered successful only if all task objectives are satisfied (e.g., successful grasp, articulated manipulation, and object placement).

\begin{table*}[t]
\centering
\caption{Overall task success rates on the \textbf{LEAP Hand}.
Results are reported as \textbf{successes / total trials} for training objects and test objects (see Fig.~\ref{fig:objects}).
Best results per column are highlighted in bold.}
\label{tab:overall_performance}
\begin{tabular}{lcccccc}
\toprule
\multirow{2}{*}{Method} &
\multicolumn{3}{c}{Training Objects} &
\multicolumn{3}{c}{Test Objects} \\
\cmidrule(lr){2-4} \cmidrule(lr){5-7}
& Grasp+Pull & Grasp+Open & Grasp+Grasp
& Grasp+Pull & Grasp+Open & Grasp+Grasp \\
\midrule
Image-based DP
& 3 / 34 & 3 / 34 & 0 / 27
& 3 / 31 & 2 / 31 & 0 / 15 \\

DP3 (raw point cloud)
& 3 / 34 & 2 / 34 & 0 / 27
& 3 / 31 & 1 / 31 & 0 / 15 \\

Object-centric DP3
& 7 / 34 & 12 / 34 & 8 / 27
& 8 / 31 & 12 / 31 & \textbf{7 / 15} \\


\midrule
\textbf{Ours (retrieval-based)}
& \textbf{22 / 34} & \textbf{23 / 34} & \textbf{12 / 27}
& \textbf{24 / 31} & \textbf{22 / 31} & 3 / 15 \\
\bottomrule
\end{tabular}
\vspace{-5pt}
\end{table*}

Across all evaluated tasks, our method consistently outperforms diffusion-based policies on both training and test objects.
The performance gap is particularly pronounced for tasks that require precise hand-object alignment and sequential coordination, such as \textbf{Grasp+Pull} and \textbf{Grasp+Open}.
These results demonstrate the effectiveness of object-centric skill segmentation and retrieval-based execution for real-world dexterous multi-task manipulation.

\paragraph{Failure Mode Analysis}
To understand performance differences, we analyze failure modes of each approach.

\textbf{Image-based DP} struggles primarily with 3D localization, particularly for small objects where inferring depth from multi-view 2D observations is challenging in low-data regimes.
Additionally, RGB-based policies exhibit sensitivity to visual domain shifts: test objects with different colors or under varying lighting conditions lead to degraded performance, consistent with prior observations~\cite{chi2023diffusionpolicy}.

\textbf{DP3 (raw point cloud)} addresses visual domain shift but introduces a different challenge: localizing target objects within cluttered raw point clouds.
Without explicit object segmentation, the policy must implicitly learn to identify and focus on task-relevant regions, which is difficult with limited demonstrations.
This failure mode aligns with recent findings in~\cite{wang2024rise,fu2025cordvip} that motivate object-centric representations.

\textbf{Object-centric DP3}, our strongest baseline, shows substantially improved performance by conditioning on segmented object point clouds.
However, it exhibits two key limitations: (i) lower spatial generalization, requiring multiple attempts to succeed across varied object placements, and (ii) higher sensitivity to the specific trained policy checkpoint, with performance varying noticeably across training runs.

\begin{figure}[t]
\center
\includegraphics[width=1\linewidth]{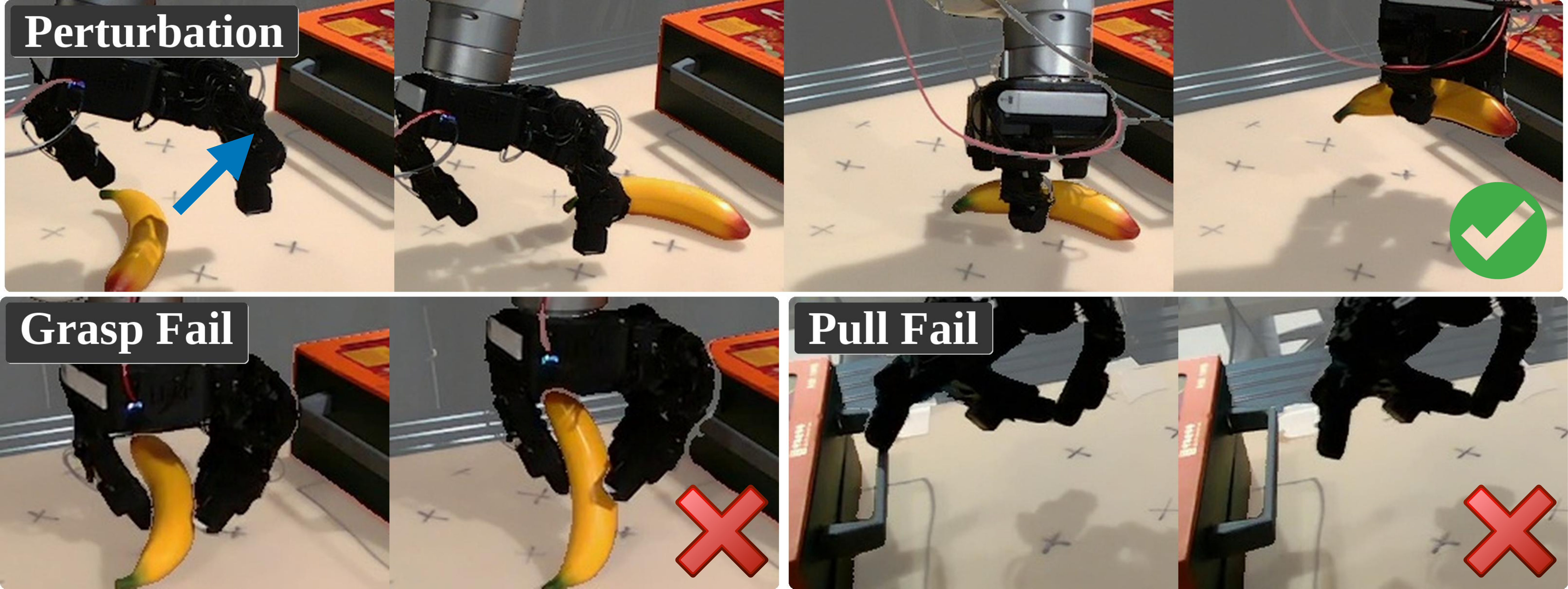}
\caption{
  Qualitative robustness and failure analysis.
  \textbf{Top row:} our method remains effective under external perturbations.
  \textbf{Bottom-left:} failure in the first task stage during banana grasping, where stable force balance is difficult.
  \textbf{Bottom-right:} failure in the pulling stage, where the hand does not contact the handle at the correct location.
}
\vspace{-10pt}
\label{fig:qualitative_result}
\end{figure}

\textbf{\method (ours)} also has failure cases, mainly in contact-sensitive stages (Fig.~\ref{fig:qualitative_result}).
The bottom-left shows a first-stage grasp failure on a banana, where stable force balance is hard.
The bottom-right shows a pulling failure where the hand does not contact the drawer handle at the correct location.

\paragraph{Safety and Robustness}
Beyond success rate, we observe a qualitative difference in failure characteristics.
All diffusion-based methods occasionally produce unsafe behaviors when encountering out-of-distribution object configurations, such as commanding the hand to collide with unexpectedly large or tall objects, triggering emergency stops.
In contrast, our retrieval-based approach exhibits more predictable failure modes: when no good demonstration match exists, the robot typically fails to establish stable contact or alignment, but does so in a controlled manner without sudden collisions.
Fig.~\ref{fig:qualitative_result} (top) further shows robustness under moderate external perturbations during execution.

\subsection{Sample Efficiency}
\label{ssec:sample_eff}

Our goal is to enable dexterous multi-task manipulation with \textbf{few real-world demonstrations}.
We therefore evaluate sample efficiency along two complementary dimensions:
\textbf{(i) demonstration quantity} and \textbf{(ii) demonstration diversity}.

\paragraph{Evaluation Protocol}
We consider a representative task (\textbf{Grasp+Open}) and evaluate performance under controlled variations in both the number and diversity of demonstrations.
All methods are trained using identical subsets of demonstrations and evaluated on held-out trials.

\begin{figure}[t]
\centering
\includegraphics[width=0.98\linewidth]{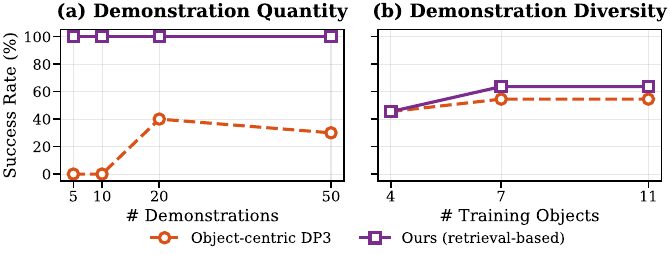}
\caption{
Sample efficiency analysis along two complementary dimensions.
\textbf{Left:} effect of demonstration quantity for a fixed object, varying the number of demonstrations in $\{5, 10, 20, 50\}$.
\textbf{Right:} effect of demonstration diversity, varying the number of distinct training objects in $\{4, 7, 11\}$ totaling 44 demonstrations.
}
\label{fig:sample_efficiency}
\vspace{-10pt}
\end{figure}

\paragraph{Effect of Demonstration Quantity}
As shown in Fig.~\ref{fig:sample_efficiency} (left), our method achieves perfect success rates with only 5 demonstrations 
per object-task configuration, maintaining 100\% across all data budgets.
In contrast, the object-centric DP3 baseline fails with 5 and 10 demonstrations and only reaches 30-40\% with 20-50 demonstrations.
This highlights that retrieval-based execution can leverage even a handful of demonstrations effectively by directly replaying object-centric skills, whereas end-to-end policies require substantially more data to learn generalizable behavior.

\paragraph{Effect of Demonstration Diversity}
Fig.~\ref{fig:sample_efficiency} (right) shows performance as the number of distinct training objects increases while fixing the total number of demonstrations.
Performance initially improves with object diversity but plateaus as the training set grows; our method consistently outperforms the baseline across all diversity levels.

These experiments reveal complementary insights: while both methods benefit from increased object diversity, our method's key advantage is \textbf{sample-efficient per-object generalization}, requiring only 5 demonstrations per object compared to 20-50 for diffusion policies.


\subsection{State Estimation and Representation Ablations}
\label{ssec:ablation}

We perform ablation studies to analyze the contribution of key design choices in our perception and execution pipeline, with a particular focus on how object state is estimated and represented for object-centric skill alignment.

\paragraph{State Estimation and Yaw Representation}

Accurate object state estimation is critical for object-centric skill alignment.
However, we hypothesize that \textit{recovering a precise 6D object pose is not a necessary condition for successful manipulation}.
In many contact-rich tasks, only a subset of state variables—such as the object centroid and a task-relevant orientation—are required for reliable execution.
To study this trade-off, we compare three state estimation variants:
\begin{itemize}
    \item \textbf{Ours:} an uncertainty-aware state estimator that tracks the object centroid and a filtered yaw angle, explicitly modeling uncertainty and avoiding over-commitment to noisy pose estimates.
    \item \textbf{FoundationPose:} an off-the-shelf 6D pose estimator~\cite{foundationposewen2024} that outputs a full object pose.
    \item \textbf{Raw PCA:} direct yaw estimation from point cloud PCA without temporal filtering or uncertainty modeling.
\end{itemize}

\begin{table}[t]
\centering
\caption{Ablation on state estimation methods.
All experiments are done using training objects.
Results are reported as \textbf{successes / total trials}.}
\label{tab:state_estimator}
\begin{tabular}{lcc}
\toprule
Estimator & Grasp+Pull & Grasp+Open \\
\midrule
FoundationPose & 12 / 34 & 9 / 34 \\
Raw PCA yaw & 11 / 34 & 12 / 34 \\
Centroid-only & 6 / 34 & 8 / 34 \\
\textbf{Ours (uncertainty-aware)} & \textbf{22 / 34} & \textbf{23 / 34} \\
\bottomrule
\end{tabular}
\vspace{-6pt}
\end{table}

As shown in Table~\ref{tab:state_estimator}, our uncertainty-aware estimator consistently outperforms both raw PCA-based yaw estimation and full 6D pose estimation.
While FoundationPose performs well on textured objects with distinctive geometry, its performance deteriorates significantly on small, symmetric, or weakly textured objects commonly encountered in our task suite.
In contrast, our approach explicitly balances estimation fidelity with robustness by tracking only task-relevant state and reasoning under uncertainty.

\paragraph{Object-Centric Canonicalization}
We further ablate the object-centric representation by comparing \textbf{centroid-only} canonicalization (ignoring orientation) against our \textbf{centroid + yaw} approach.
As shown in Table~\ref{tab:state_estimator}, removing yaw canonicalization substantially degrades performance, particularly for elongated or articulated objects such as bananas and drawer handles.
Without orientation alignment, replayed trajectories exhibit systematic misalignment that leads to failed contact initiation or unstable grasps.


\subsection{Embodiment Generalization}
\label{ssec:embodiment}

We evaluate embodiment generalization by deploying the proposed method on two dexterous hands: the 16-DoF Allegro Hand and the LEAP Hand~\cite{shaw2023leaphand}.
These hands differ in kinematic structure, joint limits, actuation bandwidth, and contact geometry.
For each embodiment, we collect the same amount of demonstration data and apply the same perception, segmentation, and skill execution pipeline.

\begin{table}[t]
\centering
\caption{Embodiment generalization across dexterous hands.
Results are reported as \textbf{successes / total trials}. ``G'' denotes grasp.}
\label{tab:embodiment}
\small
\setlength{\tabcolsep}{4pt}
\begin{tabular}{lcccc}
\toprule
\multirow{2}{*}{Method} &
\multicolumn{2}{c}{Allegro} &
\multicolumn{2}{c}{LEAP} \\
\cmidrule(lr){2-3} \cmidrule(lr){4-5}
& G+Pull & G+Open
& G+Pull & G+Open \\
\midrule
Object-centric DP3
& 17 / 34 & 2 / 34
& 7 / 34 & 12 / 34 \\

\textbf{Ours}
& \textbf{30 / 34} & \textbf{20 / 34}
& \textbf{22 / 34} & \textbf{23 / 34} \\
\bottomrule
\end{tabular}
\vspace{-6pt}
\end{table}

We compare our method against the strongest learning-based baseline under both embodiments.
Table~\ref{tab:embodiment} reports success rates on representative tasks.
Across both embodiments, our method consistently outperforms the strongest baseline.
We observe that certain tasks benefit more from specific hardware designs. For example, the Allegro Hand has higher success on \textbf{Grasp+Pull}, partially due to its stable performance on small and thin objects.
This indicates that hardware constraints can significantly influence the upper bound of achievable task performance.
Our method maintains strong performance across embodiments and consistently outperforms learning-based baselines.

\subsection{Comparison with Demonstration-Free Methods}
\label{ssec:demo_free}

A natural question is whether demonstration-free approaches, such as reinforcement learning~\cite{chen2023sequential,SOPE_2024} or grasp synthesis~\cite{wang2023dexgraspnet,li2023froggerfastrobustgrasp,liu2022differentiableforceclosure}, could replace our demonstration-based pipeline.
In practice, these methods rely on carefully engineered reward or energy functions and well-chosen initializations to elicit the desired behavior, which becomes increasingly difficult for multi-stage tasks.
We show illustrative examples of these challenges on our project website.

To promote transparency, we release all experiment videos (for all methods) on our project website (\href{https://dexmulti.github.io/}{dexmulti.github.io}).









\section{Conclusion, Limitations, and Future Work}

We introduce \method, a practical approach to concurrent prehensile and nonprehensile dexterous manipulation through object-centric skill decomposition and a retrieve-align-execute paradigm.
By decomposing demonstrations into reusable skills with well-defined temporal boundaries and combining them with uncertainty-aware state estimation, the system achieves sample-efficient learning for contact-rich multi-stage tasks.
Across three manipulation tasks and two hand embodiments, \method substantially outperforms end-to-end diffusion policy baselines while requiring far fewer demonstrations.
Our results suggest that explicit object-centric reasoning offers a practical path toward more generalizable and sample-efficient dexterous manipulation systems.

\textbf{Limitations and Future Work.}
Our approach remains dependent on high-quality real-world teleoperation demonstrations and multi-view RGB-D sensing.
Generalization is primarily limited to object geometries within the training distribution, with performance degrading under extreme configurations.
In addition, the relatively bulky hand hardware makes manipulation of very thin objects (e.g., credit cards or thin fabric) challenging.
Future work includes tactile feedback, improved hand hardware (e.g. ORCA Hand~\cite{christoph2025orcaopensourcereliablecosteffective}), and stronger single-view geometry estimation~\cite{lyu2025sam3d}.

\bibliographystyle{IEEEtran}
\bibliography{references}

@String { icra    = {IEEE International Conference on Robotics and Automation (ICRA)} }

@String { ieeera  = {IEEE Robotics and Automation Letters (RA-L)} }

@String { ijrr    = {International Journal of Robotics Research (IJRR)} }

@String { iros    = {IEEE/RSJ International Conference on Intelligent Robots and Systems (IROS)} }

@String { isrr    = {International Symposium on Robotics Research (ISRR)} }

@String { rss     = {Robotics: Science and Systems (RSS)} }

@String { tro     = {IEEE Transactions on Robotics} }

@String { iclr    = {International Conference on Learning Representations (ICLR)} }

@String { corl    = {Conference on Robot Learning (CoRL)} }

@String { cvpr    = {IEEE Conference on Computer Vision and Pattern Recognition (CVPR)} }

@String { eccv    = {European Conference on Computer Vision (ECCV)} }

@String { iccv    = {IEEE International Conference on Computer Vision (ICCV)} }

@inproceedings{li2026nonprehensile,
  title={Learning Geometry-Aware Nonprehensile Pushing and Pulling with Dexterous Hands},
  author={Yunshuang Li and Yiyang Ling and Gaurav Sukhatme and Daniel Seita},
  booktitle=icra,
  Year={2026}
}

@inproceedings{wang2023dexgraspnet,
  title={Dexgraspnet: A large-scale robotic dexterous grasp dataset for general objects based on simulation},
  author={Wang, Ruicheng and Zhang, Jialiang and Chen, Jiayi and Xu, Yinzhen and Li, Puhao and Liu, Tengyu and Wang, He},
  booktitle=icra,
  year={2023},
}

@inproceedings{chen2021multiobjectgraspingestimating,
  title={Multi-Object Grasping -- Estimating the Number of Objects in a Robotic Grasp}, 
  author={Tianze Chen and Adheesh Shenoy and Anzhelika Kolinko and Syed Shah and Yu Sun},
  booktitle=iros,
  year={2021},
}

@inproceedings{yoshikawa2001powergrasp,
  title={Optimization of Power Grasps for Multiple Objects},
  author={T. Yoshikawa and T. Watanabe and M. Daito},
  booktitle=icra,
  year={2001}
}

@inproceedings{RobotSynesthesia2024,
  title={{Robot Synesthesia: In-Hand Manipulation with Visuotactile Sensing}},
  author={Ying Yuan and Haichuan Che and Yuzhe Qin and Binghao Huang and Zhao-Heng Yin and Kang-Won Lee and Yi Wu and Soo-Chul Lim and Xiaolong Wang},
  booktitle = icra,
  year={2024},
}

@inproceedings{sun2022multi,
  title={Multi-object grasping-types and taxonomy},
  author={Sun, Yu and Amatova, Eliza and Chen, Tianze},
  booktitle=icra,
  year={2022},
}

@inproceedings{kim2025pyroki,
  title={PyRoki: A Modular Toolkit for Robot Kinematic Optimization},
  author={Kim*, Chung Min and Yi*, Brent and Choi, Hongsuk and Ma, Yi and Goldberg, Ken and Kanazawa, Angjoo},
  booktitle=iros,
  year={2025},
}

@inproceedings{he2025sequential,
  title={Sequential Multi-Object Grasping with One Dexterous Hand},
  author={Sicheng He and Zeyu Shangguan and Kuanning Wang and Yongchong Gu and Yuqian Fu and Yanwei Fu and Daniel Seita},
  booktitle=iros,
  year={2025}
}

@inproceedings{chen2024task,
  title={Task-Oriented Dexterous Hand Pose Synthesis Using Differentiable Grasp Wrench Boundary Estimator},
  author={Chen, Jiayi and Chen, Yuxing and Zhang, Jialiang and Wang, He},
  booktitle=iros,
  year={2024},
}

@inproceedings{patidar2023inhandcubereconfigurationsimplified,
  title={In-Hand Cube Reconfiguration: Simplified}, 
  author={Sumit Patidar and Adrian Sieler and Oliver Brock},
  booktitle=iros,
  year={2023},
}

@inproceedings{li2023froggerfastrobustgrasp,
  title={FRoGGeR: Fast Robust Grasp Generation via the Min-Weight Metric}, 
  author={Albert H. Li and Preston Culbertson and Joel W. Burdick and Aaron D. Ames},
  booktitle=iros,
  year={2023},
}

@inproceedings{yu2001internalforces,
  title={Computation of Grasp Internal Forces for Stably Grasping Multiple Objects},
  author = {Yong Yu and Kenro Fukuda and Showzow Tsujio},
  booktitle=iros,
  year={2001}
}

@inproceedings{harada1998kinematics,
  title={Kinematics and internal force in grasping multiple objects},
  author={Harada, Kensuke and Kaneko, Makoto},
  booktitle=iros,
  year={1998},
}

@article{dexpoint,
  title          = {DexPoint: Generalizable Point Cloud Reinforcement Learning for Sim-to-Real Dexterous Manipulation },
  author         = {Qin, Yuzhe and Huang, Binghao and Yin, Zhao-Heng and Su, Hao and Wang, Xiaolong},
  journal        = {Conference on Robot Learning (CoRL)},
  year           = {2022},
}

@inproceedings{kuang2024ramretrievalbasedaffordancetransfer,
  title={RAM: Retrieval-Based Affordance Transfer for Generalizable Zero-Shot Robotic Manipulation}, 
  author={Yuxuan Kuang and Junjie Ye and Haoran Geng and Jiageng Mao and Congyue Deng and Leonidas Guibas and He Wang and Yue Wang},
  booktitle=corl,
  year={2024},
}

@InProceedings{qi2022hand,
   author={Qi, Haozhi and Kumar, Ashish and Calandra, Roberto and Ma, Yi and Malik, Jitendra},
   title={{In-Hand Object Rotation via Rapid Motor Adaptation}},
   booktitle=corl,
   year={2022}
}

@inproceedings{lum2024gripmultifingergraspevaluation,
  title={Get a Grip: Multi-Finger Grasp Evaluation at Scale Enables Robust Sim-to-Real Transfer},
  author={Tyler Ga Wei Lum and Albert H. Li and Preston Culbertson and Krishnan Srinivasan and Aaron D. Ames and Mac Schwager and Jeannette Bohg},
  booktitle=corl,
  year={2024},
}

@inproceedings{chen2023sequential,
  title={Sequential Dexterity: Chaining Dexterous Policies for Long-Horizon Manipulation},
  author={Chen, Yuanpei and Wang, Chen and Fei-Fei, Li and Liu, C Karen},
  booktitle=corl,
  year={2023}
}

@inproceedings{fu2025cordvip,
  title={{CordViP: Correspondence-based Visuomotor Policy for Dexterous Manipulation in Real-World}}, 
  author={Yankai Fu and Qiuxuan Feng and Ning Chen and Zichen Zhou and Mengzhen Liu and Mingdong Wu and Tianxing Chen and Shanyu Rong and Jiaming Liu and Hao Dong and Shanghang Zhang},
  booktitle=rss,
  year={2025},
}

@inproceedings{wang2024dexcap,
  title = {{DexCap: Scalable and Portable Mocap Data Collection System for Dexterous Manipulation}},
  author = {Wang, Chen and Shi, Haochen and Wang, Weizhuo and Zhang, Ruohan and Fei-Fei, Li and Liu, C. Karen},
  booktitle = rss,
  year = {2024}
}

@article{wang2024rise,
  title={{RISE: 3D Perception Makes Real-World Robot Imitation Simple and Effective}},
  author={Wang, Chenxi and Fang, Hongjie and Fang, Hao-Shu and Lu, Cewu},
  journal={arXiv preprint arXiv:2404.12281},
  year={2024}
}

@inproceedings{Ze2024DP3,
  title={{3D Diffusion Policy: Generalizable Visuomotor Policy Learning via Simple 3D Representations}},
  author={Yanjie Ze and Gu Zhang and Kangning Zhang and Chenyuan Hu and Muhan Wang and Huazhe Xu},
  booktitle=rss,
  year={2024}
}

@inproceedings{chi2023diffusionpolicy,
  title={{Diffusion Policy: Visuomotor Policy Learning via Action Diffusion}}, 
  author={Cheng Chi and Zhenjia Xu and Siyuan Feng and Eric Cousineau and Yilun Du and Benjamin Burchfiel and Russ Tedrake and Shuran Song},
  booktitle=rss,
  year={2023},
}

@inproceedings{shaw2023leaphand,
  title={LEAP Hand: Low-Cost, Efficient, and Anthropomorphic Hand for Robot Learning},
  author={Shaw, Kenneth and Agarwal, Ananye and Pathak, Deepak},
  booktitle=rss,
  year={2023}
}

@inproceedings{liu2020deepdifferentiablegraspplanner,
  title={Deep Differentiable Grasp Planner for High-DOF Grippers}, 
  author={Min Liu and Zherong Pan and Kai Xu and Kanishka Ganguly and Dinesh Manocha},
  year={2020},
  booktitle=rss,
}

@inproceedings{SOPE_2024,
  title={{Learning to Singulate Objects in Packed Environments using a Dexterous Hand}},
  author={Hao Jiang and Yuhai Wang and Hanyang Zhou and Daniel Seita},
  booktitle = isrr,
  year={2024}
}

@inproceedings{shaw2025leapv2adv,
  title={{LEAP Hand V2 Advanced: Dexterous, Low-cost Hybrid Rigid-Soft Hand for Robot Learning}},
  author={Shaw, Kenneth and Pathak, Deepak},
  booktitle={IEEE-RAS International Conference on Humanoid Robots (Humanoids)},
  year={2025}
}

@misc{christoph2025orcaopensourcereliablecosteffective,
  title={ORCA: An Open-Source, Reliable, Cost-Effective, Anthropomorphic Robotic Hand for Uninterrupted Dexterous Task Learning},
  author={Clemens C. Christoph and Maximilian Eberlein and Filippos Katsimalis and Arturo Roberti and Aristotelis Sympetheros and Michel R. Vogt and Davide Liconti and Chenyu Yang and Barnabas Gavin Cangan and Ronan J. Hinchet and Robert K. Katzschmann},
  year={2025},
  eprint={2504.04259},
  archivePrefix={arXiv},
  primaryClass={cs.RO},
  url={https://arxiv.org/abs/2504.04259},
}

@misc{lyu2025sam3d,
  title={{SAM 3D: 3Dfy Anything in Images}},
  author={Lyu, Yunfan and Madapana, Naveen and Aggarwal, Vaibhav and Deng, Qi and Tremblay, Jonathan and Wang, Junwen and Zheng, Kai and Rallapalli, Rajeev and Jaipuria, Nikita and Iwana, Brian Kenji and Prasad, Ishita},
  year={2025},
  eprint={2511.16624},
  archivePrefix={arXiv},
  primaryClass={cs.CV},
  url={https://arxiv.org/abs/2511.16624}
}

@inproceedings{he2025dexgraspnet3,
  title={{DexVLG: Dexterous Vision-Language-Grasp Model at Scale}}, 
  author={Jiawei He and Danshi Li and Xinqiang Yu and Zekun Qi and Wenyao Zhang and Jiayi Chen and Zhaoxiang Zhang and Zhizheng Zhang and Li Yi and He Wang},
  booktitle=iccv,
  year={2025},
}

@inproceedings{liu2024grounding,
  title={{Grounding DINO: Marrying DINO with Grounded Pre-training for Open-set Object Detection}},
  author={Liu, Shilong and Zeng, Zhaoyang and Ren, Tianhe and Li, Feng and Zhang, Hao and Yang, Jie and Li, Chunyuan and Yang, Jianwei and Su, Hang and Zhu, Jun and others},
  booktitle=eccv,
  year={2024}
}

@inproceedings{song2020ddim,
      title={Denoising Diffusion Implicit Models},
      author={Jiaming Song and Chenlin Meng and Stefano Ermon},
      booktitle={International Conference on Learning Representations (ICLR)},
      year={2021},
}

@inproceedings{qi2017pointnet,
      title={PointNet: Deep Learning on Point Sets for 3D Classification and Segmentation},
      author={Charles R. Qi and Hao Su and Kaichun Mo and Leonidas J. Guibas},
      booktitle=cvpr,
      year={2017},
}

@inproceedings{lu2025graspahandful,
  title={Grasping a Handful: Sequential Multi-Object Dexterous Grasp Generation},
  author={Lu, Haofei and Dong, Yifei and Weng, Zehang and Pokorny, Florian T. and Lundell, Jens and Kragic, Danica},
  booktitle=ieeera,
  year={2025},
}

@inproceedings{li2024grasp,
  title={Grasp Multiple Objects with One Hand},
  author={Li, Yuyang and Liu, Bo and Geng, Yiran and Li, Puhao and Yang, Yaodong and Zhu, Yixin and Liu, Tengyu and Huang, Siyuan},
  booktitle=ieeera,
  year={2024}
}

@inproceedings{liu2022differentiableforceclosure,
   title={Synthesizing Diverse and Physically Stable Grasps With Arbitrary Hand Structures Using Differentiable Force Closure Estimator},
   author={Liu, Tengyu and Liu, Zeyu and Jiao, Ziyuan and Zhu, Yixin and Zhu, Song-Chun},
   booktitle=ieeera,
   year={2022},
}

@inproceedings{yao2023exploiting,
  title={Exploiting kinematic redundancy for robotic grasping of multiple objects},
  author={Yao, Kunpeng and Billard, Aude},
  booktitle=tro,
  year={2023},
}

@inproceedings{openai-dactyl,
  title={{Learning Dexterous In-Hand Manipulation}},
  author={OpenAI and Marcin Andrychowicz and Bowen Baker and Maciek Chociej and Rafal Jozefowicz and Bob McGrew and Jakub Pachocki and Arthur Petron and Matthias Plappert and Glenn Powell and Alex Ray and Jonas Schneider and Szymon Sidor and Josh Tobin and Peter Welinder and Lilian Weng and Wojciech Zaremba},
  booktitle = ijrr,
  year={2019}
}

@inproceedings{mason1999nonprehensile,
  title={{Progress in Nonprehensile Manipulation}},
  author={Matthew T. Mason},
  booktitle=ijrr,
  year={1999},
}

@inproceedings{rus1999inhanddexmanip,
  title={{In-Hand Dexterous Manipulation of Piecewise-Smooth 3-D Objects}},
  author={Daniela Rus},
  booktitle=ijrr,
  year={1999},
}

@article{Dreczkowski2025thousandtasksday,
   title={{Learning a thousand tasks in a day}},
   author={Dreczkowski, Kamil and Vitiello, Pietro and Vosylius, Vitalis and Johns, Edward},
   journal={Science Robotics},
   year={2025},
}

@article{eom2024mogrip,
  title = {MOGrip: Gripper for multiobject grasping in pick-and-place tasks using translational movements of fingers},
  author = {Jaemin Eom  and Sung Yol Yu  and Woongbae Kim  and Chunghoon Park  and Kristine Yoonseo Lee  and Kyu-Jin Cho},
  journal = {Science Robotics},
  year = {2024}
}

@article{yamada2015static,
  title={Static grasp stability analysis of multiple spatial objects},
  author={Yamada, Takayoshi and Yamamoto, Hidehiko},
  journal={Journal of Control Science and Engineering},
  volume={3},
  year={2015}
}

@article{ravi2024sam2,
  title={{SAM 2: Segment Anything in Images and Videos}},
  author={Ravi, Nikhila and Gabeur, Valentin and Hu, Yuan-Ting and Hu, Ronghang and Ryali, Chaitanya and Ma, Tengyu and Khedr, Haitham and R{\"a}dle, Roman and Rolland, Chloe and Gustafson, Laura and Mintun, Eric and Pan, Junting and Alwala, Kalyan Vasudev and Carion, Nicolas and Wu, Chao-Yuan and Girshick, Ross and Doll{\'a}r, Piotr and Feichtenhofer, Christoph},
  journal={arXiv preprint arXiv:2408.00714},
  year={2024}
}

@inproceedings{ester1996density,
  title={{A Density-based Algorithm for Discovering Clusters in Large Spatial Databases with Noise}},
  author={Ester, Martin and Kriegel, Hans-Peter and Sander, Jorg and Xu, Xiaowei and others},
  booktitle={International Conference on Knowledge Discovery and Data Mining (KDD)},
  year={1996}
}

@article{openai2019solvingrubikscuberobot,
  title={Solving Rubik's Cube with a Robot Hand}, 
  author={OpenAI and Ilge Akkaya and Marcin Andrychowicz and Maciek Chociej and Mateusz Litwin and Bob McGrew and Arthur Petron and Alex Paino and Matthias Plappert and Glenn Powell and Raphael Ribas and Jonas Schneider and Nikolas Tezak and Jerry Tworek and Peter Welinder and Lilian Weng and Qiming Yuan and Wojciech Zaremba and Lei Zhang},
  journal={arXiv preprint arXiv:1910.07113},
  year={2019},
}

@InProceedings{foundationposewen2024,
    author        = {Wen, Bowen and Yang, Wei and Kautz, Jan and Birchfield, Stan},
    title         = {{FoundationPose}: Unified 6D Pose Estimation and Tracking of Novel Objects},
    booktitle     = {CVPR},
    year          = {2024},
}

\clearpage
\appendix
\subsection{Additional Method Details}

\paragraph{Perception Pipeline}
We stream synchronized RGB-D images from one or more calibrated depth cameras.
Using known camera intrinsics and extrinsics, depth images are back-projected into 3D space and transformed into the robot base frame, yielding a fused scene point cloud.
The multi-view setup mitigates occlusion and improves geometric coverage during manipulation.
Given the language instruction $\ell$, we use Grounding DINO~\cite{liu2024grounding} to generate an initial 2D bounding box for each task-relevant object category.
The bounding box is then used as a prompt for SAM2~\cite{ravi2024sam2}, which performs real-time instance segmentation and tracking across frames.
Using the resulting masks, we project only the pixels corresponding to each tracked object into 3D, producing an object-only point cloud that discards background clutter (tabletop, robot arm, surrounding environment).
All points are expressed in the robot base frame to ensure consistency across views and time.

Raw segmented point clouds are often noisy and contain outliers.
To produce clean, fixed-size object point clouds, we apply the following denoising pipeline:
\begin{enumerate}
    \item \textbf{Workspace cropping:} Points are cropped to a predefined 3D bounding region around the tabletop.
    \item \textbf{Random downsampling:} The cropped cloud is uniformly downsampled to at most 8192 points to bound computation cost.
    \item \textbf{Statistical outlier removal:} Points with abnormally large distances to their neighbors are removed.
    \item \textbf{Radius outlier removal:} Sparse points with insufficient local support are discarded.
    \item \textbf{Density-based clustering:} DBSCAN~\cite{ester1996density} is applied to remove small isolated clusters, assuming segmented object point clouds form coherent structures.
    \item \textbf{Farthest point sampling (FPS):} The remaining points are downsampled to a fixed size of 512 points.
\end{enumerate}
All denoising and downsampling operations are implemented on the GPU and deployed across multiple ROS nodes.

\paragraph{Skill Segmentation}
To reason about hand-object interaction for skill segmentation, we model the dexterous hand as a point cloud.
Given a joint configuration, link-level surface points sampled from the robot URDF are transformed into the world frame via forward kinematics, producing a hand point cloud synchronized with the object point cloud.
Using these paired point clouds, we compute a soft contact signal measuring geometric proximity between the hand and each object surface~\cite{fu2025cordvip}: for each object surface point, we evaluate its distance to the hand and convert this into a continuous contact indicator.
The resulting signal provides a temporally consistent measure of interaction strength that is robust to sensor noise and partial occlusion.
Keyframes are detected as the first timestep at which sustained hand-object contact occurs for each object $o_k$, and the demonstration is segmented at these boundaries into object-centric skills $(\sigma_{i,1}, \ldots, \sigma_{i,K})$.

\paragraph{EKF State Representation}
For each object, the estimator maintains the state $\mathbf{x}_t = (\mathbf{c}_t, \psi_t, \dot{\mathbf{c}}_t, \dot{\psi}_t)$ with a Gaussian belief $p(\mathbf{x}_t) = \mathcal{N}(\boldsymbol{\mu}_t, \mathbf{P}_t)$ using constant-velocity dynamics.
The yaw observability score is computed from geometric cues (anisotropy and spatial extent of the point cloud) and the filter's yaw uncertainty.
This score is temporally smoothed and thresholded with hysteresis to produce the binary flag $\mathbb{I}^{obs}_t$.
When yaw is observable, the task-level yaw $\psi^{task}_t$ smoothly tracks the estimated yaw; when unobservable, it is held constant.

\paragraph{Yaw Selection Rule During Alignment}
Let $\psi^{retr}$ denote the yaw used during retrieval (either the task-level yaw when geometrically identifiable, or the relative alignment $\theta^*$ when not), and let $\mathbb{I}^{retr} \in \{0,1\}$ indicate whether yaw was observable at retrieval time.
At each alignment step, the yaw used for world-frame transformation is:
\begin{equation}
\psi_t =
\begin{cases}
\psi^{task}_t,
& \text{if } \mathbb{I}^{retr}=1 \text{ and } \mathbb{I}^{obs}_t=1, \\[4pt]
\psi^{task}_{t-1},
& \text{if } \mathbb{I}^{retr}=1 \text{ and } \mathbb{I}^{obs}_t=0, \\[4pt]
\psi^{retr},
& \text{if } \mathbb{I}^{retr}=0.
\end{cases}
\end{equation}
Retrieval-time observability captures ambiguity from object geometry, while real-time observability reflects estimator confidence that can vary due to occlusion.
Separating these two notions avoids spurious orientation changes under ambiguity while still exploiting reliable yaw updates when available.

\paragraph{Collision Representation for Motion Planning}
To enable real-time collision avoidance during closed-loop alignment, we approximate each obstacle point cloud with a compact multi-sphere envelope.
Given an object point cloud, we perform PCA to identify its principal axis and partition the cloud into $K$ bins along this axis (typically $K{=}6$).
For each bin, we compute a medial-axis sphere center from the local point distribution and estimate the sphere radius from orthogonal thickness (using a conservative percentile).
The resulting $K$ sphere centers and radii are temporally smoothed using critically damped second-order filters to prevent jitter.
These collision spheres, together with static workspace boundaries, are passed to the Pyroki motion planner at each control step, enabling efficient collision checking against perception-derived obstacle geometries.

\subsection{Baseline Implementation Details}
\label{app:baselines}

All baselines are trained on the same set of expert demonstrations and share the same perception pipeline where applicable.

\paragraph{Image-based DP}
We use the Diffusion Policy implementation of Chi et al.~\cite{chi2023diffusionpolicy} with a ResNet-18 visual encoder.
The policy receives RGB images from both cameras and the robot's proprioceptive state (end-effector pose and hand joint angles).
We train for 600 epochs with a prediction horizon of 32, an observation window of 3 steps, and an action chunk size of 16.

\paragraph{DP3 (raw point cloud)}
We use the 3D Diffusion Policy of Ze et al.~\cite{Ze2024DP3}, which conditions on a single scene-level point cloud synthesized from multi-view RGB-D.
The point cloud is cropped to the workspace bounding box and downsampled to 4096 points via farthest point sampling (FPS).
We train for 300 epochs with a prediction horizon of 32, an observation window of 3 steps, and an action chunk size of 16.

\paragraph{Object-centric DP3}
This baseline builds on the 3D Diffusion Policy, incorporating key ideas from CordViP~\cite{fu2025cordvip} to process object-centric point clouds instead of raw scene point clouds.
We describe the design choices below.

\textit{Point cloud input.}
Rather than encoding a single monolithic scene point cloud, we use our language-conditioned segmentation pipeline (\S\ref{ssec:perception}) to produce separate per-object point clouds.
Each object's point cloud consists of 512 points with 4 channels: XYZ coordinates plus a per-point contact map computed following CordViP~\cite{fu2025cordvip}.

\textit{Encoder architecture.}
Each point cloud is independently processed by a shared PointNet~\cite{qi2017pointnet} encoder consisting of three fully connected layers with LayerNorm and ReLU activations, producing a 256-dimensional feature vector per object.
Feature vectors from all objects are concatenated to form the visual observation embedding.
In parallel, the robot's proprioceptive state---end-effector position (3D), end-effector orientation (9D flattened rotation matrix), and hand joint angles (16D)---is encoded via a separate MLP into a 64-dimensional vector.
The point cloud and proprioceptive features are concatenated to form the full observation conditioning vector.

\textit{Diffusion policy head.}
We adopt the same conditional UNet1D architecture as DP3, with FiLM-based conditioning on the observation features.
The policy jointly predicts a 28-dimensional action: 3D TCP position, 9D rotation matrix (projected to $\mathrm{SO}(3)$ via SVD), and 16D hand joint targets.
We use 50 diffusion timesteps during training and accelerate inference with DDIM~\cite{song2020ddim} using 10 denoising steps.

\textit{Data augmentation.}
We apply SE(3) augmentation to the point clouds during training, consisting of small random translations (scale $0.01$\,m) and point-level jitter (scale $0.01$\,m), applied consistently across all point clouds in a given sample to preserve spatial relationships.

\textit{Distinction from CordViP.}
This baseline adapts CordViP's core ideas of using separately segmented object point clouds and per-point contact maps as policy input.
Following CordViP, we generate hand point clouds via a forward kinematics model and use them to compute per-point contact maps on each object surface.
However, we omit several components of the full CordViP pipeline:
(1) we do not use FoundationPose for 6D pose-driven point cloud tracking---point clouds come directly from RGB-D segmentation;
and (2) we do not employ CordViP's hand-arm coordination pre-training objectives.
This design isolates the contribution of object-centric 3D representations and contact maps from the auxiliary pre-training losses.

\textit{Training details.}
We train for 300 epochs using the AdamW optimizer with a learning rate of $1 \times 10^{-4}$, a prediction horizon of 32, an observation window of 3 steps, and an action chunk of 16.
Images, point cloud coordinates, TCP positions, and hand joint positions are linearly normalized to $[-1, 1]$ per dimension for training stability; orientations (rotation matrices or quaternions) are left unnormalized.

\subsection{Additional Experiment Details}

\paragraph{Control Frequencies and Command Interfaces}
For the arm, the action space is defined in task space as target TCP poses.
Arm commands are issued at 10\,Hz and linearly interpolated to the robot's internal control frequency of 20\,Hz.
To ensure smooth and safe execution, each arm command is constrained to remain close to the current TCP pose: translational motion is limited to at most 10\,cm per second, and rotational motion to at most 25 degrees per second.

For the hand, we directly command joint positions for each degree of freedom via the underlying PD controller provided by the hardware interface.
Hand commands are issued at 10\,Hz and internally upsampled by the robot controller to the native control frequencies (40\,Hz for the LEAP Hand and 200\,Hz for the Allegro Hand).
This unified command interface allows consistent high-level control across different embodiments while preserving smooth low-level actuation.

The perception pipeline runs fully online, producing object-centric point clouds at approximately 20\,Hz.
Hand point clouds generated via forward kinematics are published at approximately 30\,Hz.
All point cloud denoising, contact map computation, and interaction signal extraction are implemented on the GPU and deployed as ROS nodes.

\paragraph{Demonstration Recording}
Expert demonstrations are recorded at 10\,Hz, matching the high-level control frequency.
Each demonstration includes synchronized robot states (arm joint positions, hand joint positions, TCP pose), multi-view RGB-D observations, object-centric point clouds, and contact maps.
For each task, we collect demonstrations using a set of 13 distinct training objects.
For standard experiments, we collect 3--4 demonstrations per object, resulting in approximately 40 demonstrations per task.
These demonstrations are used both by our retrieval-based method and to train end-to-end learning baselines.
To evaluate sample efficiency, we additionally collect up to 50 demonstrations for selected object-task combinations.
All demonstrations are initialized from randomized object configurations across the workspace.

\subsection{Detailed Limitations and Future Work}

\paragraph{Current Limitations}
Although \method is sample-efficient relative to end-to-end baselines, it still depends on high-quality teleoperation demonstrations to define reusable skill segments.
The retrieval-and-alignment mechanism can also degrade when test-time scenes lie far outside the geometry and placement distribution seen during training, especially under severe occlusion or highly cluttered layouts.
Our current perception stack relies on calibrated multi-view RGB-D sensing and accurate language-conditioned segmentation; this improves robustness but introduces deployment overhead and sensitivity to sensor placement.
On the hardware side, the hand embodiment and fingertip geometry still limit reliable manipulation of very thin or highly deformable objects.
Finally, while object-centric decomposition improves compositionality, the current pipeline assumes a fixed interaction order and does not yet reason over alternative task graphs or recover from large sequencing errors.

\paragraph{Future Work}
Several directions could extend practicality and generalization.
First, integrating tactile sensing for both contact-state estimation and policy conditioning could improve robustness during concurrent prehensile and nonprehensile interactions, where visual ambiguity is common.
Second, adopting improved hand designs with better compliance and slimmer fingertips (e.g., LEAP v2~\cite{shaw2025leapv2adv} and ORCA Hand~\cite{christoph2025orcaopensourcereliablecosteffective}) may expand the feasible object set.
Third, reducing dependence on multi-view RGB-D by leveraging stronger single-view 3D reconstruction and geometry priors~\cite{lyu2025sam3d} would simplify deployment.
Finally, combining object-centric skill libraries with higher-level planners that can branch, re-order, and recover online is a promising path toward more open-world long-horizon dexterous manipulation.

\end{document}